\documentclass[journal,review]{IEEEtran}
\usepackage{graphicx}
\usepackage{graphics}
\usepackage{subfigure}
\usepackage{amsmath,amssymb,listings,mathrsfs}
\usepackage{color}
\usepackage{url}
 \usepackage{dsfont}
 \usepackage{bm}
 \usepackage{algorithm}
 \usepackage{cite}

\makeatletter

\newtheorem{pro}{Proposition}

\newtheorem{Lem}{Lemma}

\newcommand{\Rmnum}[1]{\expandafter\@slowromancap\romannumeral#1@}

\newlength{\IntHtx}
\newcommand{\Intvalx}[2]{\mspace{-1mu}\left.\rule[0pt]{0pt}{\IntHtx}\right|_{\,#1}^{\,#2}}

\begin{document}
%
% paper title
% can use linebreaks \\ within to get better formatting as desired
\title{Variational based Mixed Noise Removal with CNN Deep Learning Regularization}
%
%
% author names and IEEE memberships
% note positions of commas and nonbreaking spaces ( ~ ) LaTeX will not break
% a structure at a ~ so this keeps an author's name from being broken across
% two lines.
% use \thanks{} to gain access to the first footnote area
% a separate \thanks must be used for each paragraph as LaTeX2e's \thanks
% was not built to handle multiple paragraphs
%

\author{Faqiang Wang, Haiyang Huang, Jun Liu
%%\thanks{This work was supported in part by
%%the National Natural Science Foundation of China (No. 11071023).}
\thanks{Faqiang Wang, Haiyang Huang and Jun Liu are with School of Mathematical Sciences, Laboratory of Mathematics and
Complex Systems, Beijing Normal University, Beijing 100875, P.R. China. Email:jliu@bnu.edu.cn}
%\thanks{Stanley Osher is with Department of Mathematics, UCLA, 520 Portola Plaza, Los Angeles, CA 90095, USA.
%Email:sjo@math.ucla.edu}
}

\maketitle

\begin{abstract}
In this paper, the traditional model based variational method and learning based algorithms are naturally integrated to address
mixed noise removal problem. To be different from single type noise (e.g. Gaussian) removal,
it is a challenge problem to accurately discriminate noise types and levels for each pixel. We propose a variational method to iteratively estimate the noise parameters, and then the algorithm can automatically classify the noise according to the different statistical parameters. The proposed variational problem can be separated into
regularization, synthesis, parameter estimation and noise classification four steps with the operator splitting scheme.
Each step is related to an optimization subproblem. To enforce the regularization, the
deep learning method is employed to learn the natural images priori. Compared with some model based regularizations, the CNN regularizer can significantly improve the
quality of the restored images. Compared with some learning based methods,
the synthesis step can produce better reconstructions by analyzing the recognized noise types and levels.
In our method, the convolution
neutral network (CNN) can be regarded as an operator which associated to a variational functional. From this viewpoint, the proposed method can be extended to many image reconstruction and inverse problems. Numerical experiments in the paper show that our method can achieve some state-of-the-art results for mixed noise removal.
\end{abstract}


% For peer review papers, you can put extra information on the cover
% page as needed:
% \ifCLASSOPTIONpeerreview
% \begin{center} \bfseries EDICS Category: 3-BBND \end{center}
% \fi
%
% For peerreview papers, this IEEEtran command inserts a page break and
% creates the second title. It will be ignored for other modes.
\IEEEpeerreviewmaketitle

\section{Introduction}
Images are always contaminated by noise during acquisition and transmission. Usually, the distributions of random noise are assumed to be some standard probabilistic distributions, such as Gaussian, Poisson, Gamma distributions and so on. The additive noise model can be easily written as
\begin{equation*}
f=u+n,
\end{equation*}
where $f,u,n$ are observed image, true image and noise, respectively.

Many methods have been proposed \cite{Aubert2002}\cite{Buades2005a} to obtain the clean image from the observed data. Model based methods are traditional and popular techniques. In which filter method is a very classical technique, some representative methods such as Gaussian filters\cite{Geusebroek2003}, Gabor filters\cite{lindenbaum1994on} and median type filters\cite{Chen2001}\cite{Ko1991}  are still very popular since their simple implementations. 
Wavelet based approaches\cite{Katkovnik2010}\cite{Dabov2009}\cite{Portilla2001} suppress the high frequency coefficients by thresholding and statistical approaches\cite{biemond1990maximum}\cite{Liu} treat noise as some realizations of a random variable. They are two powerful methods for image denoising. Variational method\cite{meyer2001}\cite{Rudin1992} is another useful and efficient tool. This approach is to minimize a cost functional which contains a data fidelity and regularization terms
\begin{equation*}
F(u)=E(u)+\lambda J(u),
\end{equation*}
where $E(u)$ is the data fidelity to measure the discrepancy between the true and the observed data. It can be derived from the maximum likelihood estimation of noise. $J(u)$ serves as regularization formulating image prior. Variational methods draw extensive researches since these methods can naturally equip model with regularization and flexibly integrate the advantages of different methods. Meanwhile, the Total Variation(TV)\cite{Rudin1992} regularization has been proven its success on denoising and inverse problems \cite{Liu2010}\cite{Lysaker2003}\cite{Bresson2008}. However, TV can not preserve the repeated tinny image details such as textures. To better capture these image details, nonlocal methods \cite{Buades2005}\cite{Gilboa2008}\cite{buades2008nonlocal} were proposed. These methods take full use of the self-similarity properties existing in an image, which can be integrated in a variational methods naturally. The nonlocal methods always have better performance than local methods on denoising. 
However, the weighting function existing in the nonlocal model are usually difficult to be determinated. 
There are many improved nonlocal methods based on the self-similarity properties among image patches, such as BM3D\cite{Dabov2007}, learned
simultaneous sparse coding(LSSC)\cite{mairal2009non-local} and weighted nuclear norm
minimization(WNNM)\cite{Gu2014}. 

Learning based methods\cite{Bouvrie2006Notes}\cite{Hinton2006A} draw much attention recently for its outstanding denoising performance. Mathematically, the learning based methods can be expressed as
\begin{equation*}
O=F(I;\Theta),
\end{equation*}
where $F$ is a nonlinear operator functioned by recursion with parameters set $\Theta$, $I$ is the input data and $O$ represents the output data. Given some data pair $(I_i,O_i)$,  the model can be trained precisely to fit the given samples. Obviously, the learning based model can be used in many fields, so long as the sample data are prepared enough and properly, such as denoising\cite{Zhang2016Beyond}, image classification\cite{Krizhevsky2013ImageNet} and other interesting application\cite{Gatys2016Image}\cite{Hinton2012Deep}. The learning especially neural network based denoising methods have been proposed in many works, and most of these works establish different kinds of networks as denoisers, such as commonly used convolutional neural network\cite{Jain2008Natural}\cite{Zhang2016Beyond}, multi-layer perceptron\cite{Burger2012Image} and stacked sparse denoising auto-encoders method\cite{Xie2012Image}.

Most of the works assume the noise was single white Gaussian noise, which can be removed by a $L^2$-based fidelity\cite{Rudin1992} term. With other noise assumptions, the data fidelity can be different, such as $L^{1}$ based fidelity for impulse noise\cite{Nikolova2004} \cite{Liu2010} (including pepper-and-salt noise and random valued noise) and point-wise based fidelity for Possion noise\cite{le2007a}. However, the noise model is more complicated in practical application since the changeable imaging environments. More precise and reasonable, the noise should be modeled by mixture distributions such as Gaussian-Gaussian, Gaussian-impulse and so on. Unfortunately, the single type noise removal models are not suitable for the mixed noise models any more\cite{Liu2009}, since there is no unified data fidelity can be used in mixed noise removal model, which makes mixed noise removal troublesome.

The key point of mixed noise removal is to determine precisely the type of noise in each pixel. Lopez-Rubio\cite{Lopez-Rubio2010} gives a kernel estimation method to remove the Gaussian-impulse noise, which is based on Bayesian classification of each pixel. Xiao et al.\cite{Xiao2011} establish a $l_{1}-l_{0}$ model for Gaussian-impulse noise removal. Liu. et al.\cite{Liu2009} propose an adaptive mixed noise removal models based on EM process, which demonstrates good performance.

Though learning based methods show better denoising performance under single type noise such as Gaussian noise, these kinds of methods need large amount of labeled samples, which limits the application and development of these methods on mixed noise removal. As for variational based methods, most of the variational based methods needs only one image, and these kinds of methods can integrate the prior (regularization) of image flexibly. However, the prior existing in a variational model is always based on low level image information and single image, which is always unsatisfied.

In this paper, we extend our previous EM based mixed noise removal method \cite{Liu2013}, and integrate a CNN process as regularization to propose a new variational method. In our method, the variational process can estimate the noise parameters iteratively and it can be used  to classify noise types and levels in each pixel. By splitting methods, we can separate our algorithm into four steps: regularization, synthesis, parameter estimation and noise classification, in which each step is related a minimization problem and can be optimized efficiently. Meanwhile, we employ the deep learning method (CNN) to learn the natural images priori in our algorithm, which strengthens the regularization priori.

The CNN based regularization can better catch the image prior, image synthesis will correct the over-smooth effect by CNN process, and noise estimation can give CNN denoiser a good noise estimation to behave well, all these step work together and one can get some satisfactory restored results. To the best of knowledge, this is the first attempt integrating the variational mixed noise removal methods and learning based methods together.

The rest of this paper are organized as follows: we review the related work in section II, the proposed model and details of algorithm are presented in section III, numerical experiments of the proposed model are given in section IV. We give the conclusion and further research in section V.

\section{Related Work}
In this paper, we consider the additive mixed noise removal.
To be different from most of denoising works, here the mixed noise is assumed to be 
\begin{equation}\label{NoiseModel}
n=\left\{
\begin{array}{ll}
n_{1},~with~probability~r_1,\\
n_{2},~with~probability~r_2,\\
\cdots,\cdots\\
n_{K},~with~probability~r_K,\\
\end{array}
\right.
\end{equation}
where $n_k$ is the $k$-th noise component with probability density function (PDF) $p_k$, $r_k$ is a unknown mixture ratio and satisfies $\sum_{k=1}^Kr_k=1.$

The proposed method is built upon our previous work \cite{Liu2013}.
Let us first review some results in \cite{Liu2013}.
\begin{Lem}[\cite{Liu2013}]\label{lem1}
The PDF of mixed noise model (\ref{NoiseModel}) is
  $$p(z)=\sum_{k=1}^{K}r_k p_k(z).$$
\end{Lem}

Once the PDF is given, then by assuming the noise is independent identically distributed and
one can get its negative log-likelihood functional as follows:
\begin{equation*}
  \mathcal{L}(u,\bm\Theta)=-\sum_{x\in\Omega}\ln\sum_{k=1}^K r_k p_k(u(x)-f(x)).
\end{equation*}
Here $\bm\Theta$ is a statistical parameter set contains
noise parameters such as mixture ratios, means and
variances.

Usually, this likelihood functional can be chosen as the fidelity term in variational method.
However, to be different from the single Gaussian noise case, here $\mathcal{L}$ is not quadratic and it is not easy to be efficiently optimized. One alternative way is to minimize a simple upper bound functional of $\mathcal{L}$. In \cite{Liu2013}, such a upper bound functional named
$\mathcal{H}$ had been found as
\begin{equation}\label{Hformula}
\begin{array}{rl}
  \mathcal{H}(u,\bm \Theta,\mathbf{w})=& -\displaystyle\biggl.\sum_{x\in\Omega}
\sum_{k=1}^{K}\ln [r_kp_k(u(x)-f(x))]w_k(x)\\
&+\displaystyle\biggl.\sum_{x\in\Omega}\sum_{k=1}^{K}w_k(x)\ln w_k(x).
\end{array}
\end{equation}
Here $\bm\Theta=(r_{1},...,r_{K},\sigma_{1},...,\sigma_{K})$ is the parameters set, and $\mathbf{w}: \Omega\rightarrow (0,1)^K$ is a vector-valued function with its $k$-th component function as $w_k$. Moreover, $\mathbf{w}$ must satisfy a segmentation condition that
$$\mathbf{w}\in\mathbb{S}=\{\mathbf{w}(x): 0<w_k(x)<1, \text{and} \sum_{k=1}^{K}w_k(x)=1, \forall x\in\Omega\}.$$

In \cite{Liu2013}, it has been shown that the three variables functional $\mathcal{H}$ is a upper bound of $\mathcal{L}$, i.e.
\begin{Lem}[Commutativity of log-sum, \cite{Liu2013}]\label{lem2}
\begin{equation*}
\mathcal{L}(u,\bm\Theta)=\underset{\mathbf{w}\in\mathbb{S}}{\min}
~~\mathcal{H}(u,\bm\Theta,\mathbf{w}).
\end{equation*}
\end{Lem}

It seems that $\mathcal{H}$ is more complicated than $\mathcal{L}$ since there is an extra variable $\mathbf{w}$ in
$\mathcal{H}$. However, to minimize $\mathcal{H}$ is easier since the $\mathcal{H}$-problem would become quadratic with respect to $u$, and $\Theta$ always would have a closed-form solution in some cases. Moreover, the introduced $\mathbf{w}$ is a probability which indicates
the noise at each pixel comes from which mixture component, and thus the noise would be classified by $\mathbf{w}$ according to different statistical parameters.

To optimize $\mathcal{H}$, the alternating minimization could be employed, and one can get
the following iteration scheme:
\begin{equation}\label{AMu_f_theta}
\left\{
  \begin{array}{rl}
  \mathbf{w}^{\nu+1}&=\underset{\mathbf{w}\in \mathbb{S}}{\arg\min}~~\mathcal{H}(u^\nu,\bm\Theta^{\nu},\mathbf{w}),\\
 (u^\nu,\bm\Theta^{\nu+1})&=\underset{u,\bm\Theta}{\arg\min}~~\mathcal{H}(u,\bm\Theta,\mathbf{w}^{\nu+1}).\\
  \end{array}
  \right.
\end{equation}

For such a scheme, it has been shown that
\begin{Lem}[Energy Descent, \cite{Liu2013}]\label{Lem3}
The sequence $(u^\nu,\bm\Theta^{\nu})$ produced by iteration scheme (\ref{AMu_f_theta}) satisfies
  \begin{equation*}
    -\mathcal{L}(u^{\nu+1},\Theta^{\nu+1})\leqslant-\mathcal{L}(u^\nu,\Theta^{\nu}).
  \end{equation*}
\end{Lem}

According to this lemma, to optimize $\mathcal{L}$ can be replaced by $\mathcal{H}$ by adding a variable $\mathbf{w}$ and log-sum interchange. Based on this fact, the authors in \cite{Liu2013} proposed a variational model with dictionary learning to denoise a variety of mixed noise such as Gaussian-Gaussian, impulse and Gaussian-Impulse mixtures. However,
such a dictionary learning is images driven and thus the learning and denoisng procedures are
synchronized. In addition, it is hardly to split them into two separated tasks, and thus the algorithm would be very time-consuming. One more thing, the dictionary learning is linear and low level learning method and it could not find some nonlinear and deep image priori in natural images. In this paper, we will integrate the deep learning method to improve it.

\section{The Proposed Method}
In this section, we will built a general variational model with CNN regularization for mixed noise removal.
\subsection{General Model}
%We propose the following saddle point problem for general
%mixed noise removal:
%\begin{equation*}
%\underset{u,v}{\min}\underset{\mu}{\max}
%\left\{
%\begin{array}{l}
%\mathcal{H}(v,\bm \Theta,\mathbf{w} )+<\mu,u-v>+\frac{\eta}{2}||u-v||^2_2\\
%+\lambda_1 \mathcal{J}(u)+\lambda_2 \text{TV} (v)
%\end{array}
%\right\}.
%\end{equation*}

The general mixed noise removal model could be
\begin{equation}\label{generalE}
\begin{array}{l}
(u^*,\bm \Theta^*,\mathbf{w}^*)=\underset{u,\bm \Theta,\mathbf{w}\in\mathbb{S}}{\arg\min}\left\{
\begin{array}{l}
%\mathcal{E}(u,\bm \Theta,\mathbf{w})\\
\mathcal{H}(u,\bm \Theta,\mathbf{w})+\lambda_1 \mathcal{J}(u)
\end{array}
\right\},
\end{array}
\end{equation}
where $\mathcal{H}$ is defined in (\ref{Hformula}) and $\mathcal{J}$ is a learning-based regularization term,
$\lambda_1>0$ is a balance parameter which control the smoothness of the restorations.

By applying the well-known alternating minimization scheme, we can get

\begin{subequations}\label{eq:ADMM_0}
\begin{align}
u^{\nu+1}=&\underset{u}{\arg\min}~~\lambda_1 \mathcal{J}(u)
+\mathcal{H}(u,\bm \Theta^{\nu},\mathbf{w}^{\nu}), \label{eq:u_ori} \\ %\notag
\bm\Theta^{\nu+1}=&\underset{\bm\Theta}{\arg\min}~~\mathcal{H}(u^{\nu+1},\bm \Theta,\mathbf{w}^{\nu+1}), \label{eq:theta_ori} \\
\mathbf{w}^{\nu+1}=&\underset{\mathbf{w}\in\mathbb{S}}{\arg\min}~~\mathcal{H}(u^{\nu+1},\bm \Theta^{\nu+1},\mathbf{w}).\label{eq:w_ori}
\end{align}
\end{subequations}

In order to use CNN, we must split the optimization problem (\ref{eq:u_ori}). Let us introduce a auxiliary function $v$ and reform the above problem as
\begin{equation*}
\begin{array}{l}
\underset{u,v}{\min}\left\{\mathcal{H}(v,\bm \Theta^{\nu},\mathbf{w}^{\nu})+\lambda_1 \mathcal{J}(u)\right\}~~ s.t~~ u=v.
\end{array}
\end{equation*}

Then by applying the well-known augmented Lagrangian method (ALM) \cite{Tai2009}, one can get a saddle problem
\begin{equation*}
\underset{u,v}{\min}~\underset{\mu}{\max}
\left\{
\begin{array}{l}
\mathcal{H}(v,\bm \Theta^{\nu},\mathbf{w}^{\nu})+<\mu,u-v>+\frac{\eta}{2}||u-v||^2_2\\
+\lambda_1 \mathcal{J}(u)
\end{array}
\right\}.
\end{equation*}

We notice that the above functional with respect to $v$ is an image synthesis process and we can add a TV regularizer for $v$ to reduce some artificial effect such as blurs cased by averaging. Meanwhile, the introduction of TV can be seem as a generalization of our algorithm, since the parameter $\lambda_{2}$ can be set to 0 which equals to the original question. Thus we get
\begin{equation*}
\underset{u,v}{\min}~\underset{\mu}{\max}
\left\{
\begin{array}{l}
\mathcal{H}(v,\bm \Theta^{\nu},\mathbf{w}^{\nu})+<\mu,u-v>+\frac{\eta}{2}||u-v||^2_2\\
+\lambda_1 \mathcal{J}(u)+\lambda_2 \text{TV} (v)
\end{array}
\right\}.
\end{equation*}

It produces the standard ALM iteration scheme
\begin{equation*}
\begin{array}{rl}
(u^{\nu+1},v^{\nu+1})=&\underset{u,v}{\arg\min} \left\{
\begin{array}{l}
\frac{\eta}{2}||u-v-\mu^{\nu}||^2_2+\lambda_1 \mathcal{J}(u)
\\
+\mathcal{H}(v,\bm \Theta^{\nu},\mathbf{w}^{\nu})+\lambda_2 \text{TV}(v)
\end{array}
\right\},
\\
\mu^{\nu+1}=&\mu^{\nu}+\epsilon(v^{\nu+1}-u^{\nu+1}).\\
\end{array}
\end{equation*}
where $\epsilon$ serves as step size.

By applying the well-known alternating minimization scheme, together with
(\ref{eq:theta_ori}) and (\ref{eq:w_ori}), we can get

\begin{subequations}\label{eq:ADMM}
\begin{align}
u^{\nu+1}=&\underset{u}{\arg\min}~~\lambda_1 \mathcal{J}(u)
+\frac{\eta}{2}||u-v^{\nu}-\mu^{\nu}||^2_2, \label{eq:u_pro} \\ %\notag
v^{\nu+1}=&\underset{v}{\arg\min}~~\left\{
\begin{array}{l}
\mathcal{H}(v,\bm \Theta^{\nu},\mathbf{w}^{\nu})
+\lambda_2\text{TV}(v)\\
+\frac{\eta}{2}||u^{\nu+1}-v-\mu^{\nu}||^2_2
\end{array}
\right\},\label{eq:v_pro}
\\
\mu^{\nu+1}=&\mu^{\nu}+\epsilon(v^{\nu+1}-u^{\nu+1}),\label{eq:mu_pro}\\%\notag\\
\bm\Theta^{\nu+1}=&\underset{\bm\Theta}{\arg\min}~~\mathcal{H}(u^{\nu+1},\bm \Theta,\mathbf{w}^{\nu}), \label{eq:theta_pro} \\
\mathbf{w}^{\nu+1}=&\underset{\mathbf{w}\in\mathbb{S}}{\arg\min}~~\mathcal{H}(u^{\nu+1},\bm \Theta^{\nu+1},\mathbf{w}),\label{eq:w_pro}
\end{align}
\end{subequations}

The above 5 subproblems implies that we can split the mixed noise removal problem into
Gaussian noise removal (renewing $u$), fidelity term choice (renewing $v$), noise putback (Lagrangian multiplier $\mu$ updating), noise parameters estimation ($\bm\Theta$ updating), noise classification ($\mathbf{w}$ updating) 5 steps. In the next, we will show how to solve each
subproblem.\

Let us mention our model (\ref{generalE}) can handle many types mixture noise such as Gaussian-Gaussian mixed noise with different standard deviations, impulse noise (salt and pepper noise and random value noise), Gaussian-impulse mixed noise and so on. One just need to chose different $p_k$ to finish different types mixed noise removal problem. For Gaussian-Gaussian mixture, $\mathcal{H}$ would be quadratic with respect to $v$ and all of these five problem would be easily solvable. \\
\textbf{$u$ problem:}\\
$u$ problem is a standard Gaussian noise removal problem. In order to enforce the image priori, we can employ the popular deep learning methods such as CNN as regularizer\cite{Zhang2017learning}.
Suppose the property of $\mathcal{J}$ is good enough such as differentiable, then $u^{\nu+1}$ must satisfy
\begin{equation}\label{cnn}u^{\nu+1}=v^{\nu}+\mu^{\nu}-\mathscr{L}(u^{\nu+1}),\end{equation}
where $\mathscr{L}$ is an operator with $\mathscr{L}(u^{\nu+1})=\frac{\lambda_1}{\eta}\frac{\delta \mathcal{J}}{\delta u}\Intvalx{u=u^{\nu+1}}{}$.
Ideally, $\mathscr{L}(u^{\nu+1})$ should be the noise $n$ for a additive noise removal problem.
Thus we can use CNN to learn Gaussian noise for different kinds of levels in variety of natural images. In this sense, we can regard the CNN as a variational of a functional. In PDE denoising method, a very simple example of $\mathscr{L}$ is the negative Laplace operator, i.e $\mathscr{L}=-\triangle$, which can be regarded as a trained single layer CNN with isotropic diffusion convolution kernel. In such a case, we can easily get the functional $\mathcal{J}(u)=\frac{\eta}{2\lambda_1}||\nabla u||^2_2$.
 Though such a simple CNN is not good enough to preserve the image edges well, it can enlighten us to use some more complicated CNN with multi-layers and nonlinear kernels. In the general cases, we can not get the related closed-form functional $\mathcal{J}$ for CNN operator $\mathscr{L}$.

 Therefore, we can simulate white Gaussian noise with different variances and put them into all kinds of natural images to produce plenty of train samples. The powerful learning ability of CNN ensure that the trained CNN can distinguish the different levels of white Gaussian noise. Once the noise in the images is identified, then the clean image can be easily recovered.

Instead of solving a linear or nonlinear PDE in traditional variational method, here we employ a CNN to find noise. Numerical experiments show that this step can greatly improve the quality of the restorations.\\
There are many learning base method to remove Gaussian noise such as \cite{Xie2012Image}\cite{Zhang2017learning}.
In this paper, we choose the recent CNN based denoiser\cite{Zhang2017learning}.
 
\textbf{$v$ problem:}\\
This subproblem would lead to a TV system and can be efficiently solved by many TV solvers, such as Chambolle dual method\cite{Chambolle2004}, primary dual method\cite{Chan1999}\cite{esser2010a}\cite{zhang2011a}, split Bergman method \cite{Goldstein2009a} \cite{Getreuer2012Rudin}, augmented lagrangian method \cite{Tai2009}. For the Gaussian mixture noise,
it is the ROF model. Here the weight $\mathbf{w}$ can ensure the model assigns the different fidelity terms to the pixels contaminated by different levels or types noise. This procedure can greatly improve the quality of restorations.\\
\textbf{$\bm\Theta,\mathbf{w}$ problems:}\\
These two subproblems are exactly an EM process. For a given noise $f-u^{\nu+1}$, these two steps can give an estimation of noise variances and classify the noise into different classes according to the estimated noise parameters.

Some detailed noise distribution are discussed below, including Gaussian-Gaussian type noise, impulse noise and Gaussian-impulse noise.

\subsection{Gaussian Mixture Model }
%For simplification, we just list 2-component Gaussian mixture noise.
 Assume $p_1,p_2,\cdots,p_K$ be Gaussian functions with different variances $\sigma_1^2,\sigma_2^2,\cdots,\sigma_K^2$, respectively, i.e.
\begin{equation*}
p_k(z)=\frac{1}{\sqrt{2\pi}\sigma_k}\exp(-\frac{z^2}{2\sigma_k^2}), k=1,2,\cdots,K,
\end{equation*}
then ignoring some constant terms, we can define $\mathcal{H}$ as
\begin{equation}\label{Hformula_s1}
\begin{array}{ll}
  \mathcal{H}(u,\bm r,\bm\sigma^2,\mathbf{w})\\
  =\displaystyle\biggl.\frac{1}{2}\sum_{x\in\Omega}
\sum_{k=1}^{K}\frac{\left(u(x)-f(x)\right)^2}{\sigma_k^2}w_k(x)
-\displaystyle\biggl.\sum_{x\in\Omega}\sum_{k=1}^{K}w_k(x)\ln r_k\\
+\displaystyle\biggl.\frac{1}{2}\sum_{x\in\Omega}\sum_{k=1}^{K}w_k(x)\ln \sigma_k^2
+\displaystyle\biggl.\sum_{x\in\Omega}\sum_{k=1}^{K}w_k(x)\ln w_k(x),
\end{array}
\end{equation}
where $\bm r=(r_1,r_2,\cdots,r_K),\bm \sigma^2=(\sigma_1^2,\sigma_2^2,\dots,\sigma_K^2)$

Here we give the Gaussian mixed noise removal model:
\begin{equation}\label{model1}
\begin{array}{l}
(u^*,\bm \Theta^*,\mathbf{w}^*)=\underset{u,\bm \Theta,\mathbf{w}\in\mathbb{S}}{\arg\min}\left\{
\begin{array}{l}
%\mathcal{E}(u,\bm \Theta,\mathbf{w})\\
\mathcal{H}(u,\bm \Theta,\mathbf{w})+\lambda_1 \mathcal{J}(u)
\end{array}
\right\},
\end{array}
\end{equation}

So details of iteration scheme (\ref{eq:ADMM}) becomes:\\
%
%\begin{subequations}\label{eq:gaussian-gaussian-ADMM}
%\begin{align}
%u^{\nu+1}=&\underset{u}{\arg\min}~~\lambda_1 \mathcal{J}(u)
%+\frac{\eta}{2}||u-v^{\nu}-\mu^{\nu}||^2_2, \label{eq:u_pro1} \\ %\notag
%v^{\nu+1}=&\underset{v}{\arg\min}~~\left\{
%\begin{array}{l}
%\mathcal{H}(v,\bm \Theta^{\nu},\mathbf{w}^{\nu})
%+\lambda_2\text{TV}(v)\\
%+\frac{\eta}{2}||u^{\nu+1}-v-\mu^{\nu}||^2_2
%\end{array}
%\right\},\label{eq:v_pro1}
%\\
%\mu^{\nu+1}=&\mu^{\nu}+\epsilon(v^{\nu+1}-u^{\nu+1}),\label{eq:nu1}\\
%\bm\Theta^{\nu+1}=&\underset{\bm\Theta}{\arg\min}~~\mathcal{H}(u^{\nu+1},\bm \Theta,\mathbf{w}^{\nu}), \label{eq:theta_pro1} \\
%\mathbf{w}^{\nu+1}=&\underset{\mathbf{w}\in\mathbb{S}}{\arg\min}~~\mathcal{H}(u^{\nu+1},\bm \Theta^{\nu+1},\mathbf{w}),\label{eq:w_pro1}
%\end{align}
%\end{subequations}
%
\textbf{$v$ problem (\ref{eq:v_pro}):} \\
\begin{equation*}
v^{\nu+1}=\underset{v}{\arg\min}~~\left\{
\begin{array}{l}
\frac{1}{2}\sum\limits_{x\in\Omega}
\sum\limits_{k=1}^{K}\frac{\left(v(x)-f(x)\right)^2}{(\sigma_k^2)^{\nu}}w^{\nu}_k(x)+\\
\lambda_2\sum\limits_{x\in\Omega}||\nabla v(x)||
+\frac{\eta}{2}||u^{\nu+1}-v-\mu^{\nu}||^2_2.
\end{array}
\right\}
\end{equation*}

Since the existence of TV term, we can adopt some splitting methods, such as split Bergman method \cite{Goldstein2009a} \cite{Getreuer2012Rudin} and augmented lagrangian method \cite{Tai2009}. Here we introduce an auxiliary variable $d=\nabla v$ and give the iteration by split Bregman

\begin{subequations}\label{eq:gaussian-gaussian-ADMM2}
\begin{align}
(v^{\nu+1},d^{\nu+1})=&\underset{v,d}{\arg\min}\left\{
\begin{array}{l}
\frac{1}{2}\sum\limits_{x\in\Omega}
\sum\limits_{k=1}^{K}\frac{\left(v(x)-f(x)\right)^2}{(\sigma_k^2)^{\nu}}w^{\nu}_k(x)\\
+\lambda_2\sum\limits_{x\in\Omega}||d(x)||\\
+\frac{\lambda}{2}||d-\nabla v - b||^{2}_{2}\\
+\frac{\eta}{2}||u^{\nu+1}-v-\mu^{\nu}||^2_2,
\end{array}
\right\}\label{eq:v_pro2}
\\
b^{\nu+1}=&b^{\nu}+\lambda(\nabla v^{\nu+1}-d^{\nu+1}),\notag\\
\end{align}
\end{subequations}

Furthermore,
\begin{equation*}
v^{\nu+1}=\underset{v}{\arg\min}\left\{
\begin{array}{l}
\displaystyle\biggl.\frac{1}{2}\displaystyle\biggl.\sum\limits_{x\in\Omega}
\displaystyle\biggl.\sum\limits_{k=1}^{K}\frac{\left(v(x)-f(x)\right)^2}{(\sigma_k^2)^{\nu}}w^{\nu}_k(x)\\
+\displaystyle\biggl.\frac{\lambda}{2}||d-\nabla v - b||^{2}_{2}\\
+\displaystyle\biggl.\frac{\eta}{2}||u^{\nu+1}-v-\mu^{\nu}||^2_2,
\end{array}
\right\}\label{eq:v_pro2}
\end{equation*}
can be solved by first-order optimal condition, which equals to solve a linear system:
\begin{equation}\label{up_v}
\begin{array}{ll}
&\displaystyle\biggl.[\sum\limits_{k=1}^{K}\frac{w^{\nu}_{k}(x)}{(\sigma^{2}_{k})^{\nu}}-\lambda\Delta+\eta\displaystyle\biggl.]v(x) \\ =&\displaystyle\biggl.[\sum\limits_{k=1}^{K}\frac{w^{\nu}_{k}(x)}{(\sigma^{2}_{k})^{\nu}}\displaystyle\biggl.]f(x)+\lambda div(b^{\nu}(x)-d^{\nu}(x))+\eta(u^{\nu+1}(x)-\mu^{\nu}(x)),
\end{array}
\end{equation}
and
\begin{equation*}
{d^{\nu+1}}=\arg\min\limits_{d}\left\{\lambda_{2}\sum\limits_{x\in\Omega}||d(x)||+\frac{\lambda}{2}||d-\nabla v^{\nu+1}-b^{\nu}||^{2}_{2}\right\},
\end{equation*}
can be solved by shrinkage operator\cite{Beck2009A}:
\begin{equation}\label{up_d}
d^{\nu+1}(x)=\frac{\nabla v^{\nu+1}(x)+b^{\nu}(x)}{||\nabla v^{\nu+1}(x)+b^{\nu}(x)||}\max\{||\nabla v^{\nu+1}(x)+b^{\nu}(x)||-\frac{\lambda_{2}}{\lambda},0\}.
\end{equation}
\textbf{$\bm\Theta$ problem (\ref{eq:theta_pro}):}\\
 The parameter set $\bm\Theta$ consists of {$\bm r$}=$(r_{1},...,r_{K})$ and {$\bm\sigma$}=$(\sigma_{1},...,\sigma_{K})$, they would have closed-form solutions.
The minimization problem with respect to $\bm r$ and $\bm \sigma$ can be written as
\begin{subequations}\label{eq:paras}
\begin{align}
\bm r^{\nu+1}=&\arg\min\limits_{\bm r}\left\{
-\displaystyle\biggl.\sum_{x\in\Omega}\sum_{k=1}^{K}w^{\nu}_k(x)\ln r_k\right\},\label{eq:alpha}
\\
\bm\sigma^{\nu+1}=&\arg\min\limits_{\bm\sigma}\displaystyle\biggl.\left\{\frac{1}{2}\sum_{x\in\Omega}
\sum_{k=1}^{K}\frac{\left(u(x)-f(x)\right)^2}{\sigma_k^2}w^{\nu}_k(x)\right.\notag\\
&\left.+\displaystyle\biggl.\frac{1}{2}\sum_{x\in\Omega}\sum_{k=1}^{K}w^{\nu}_k(x)\ln \sigma_k^2\right\},\label{eq:sigma}
\end{align}
\end{subequations}

With the weights constrain $\sum\limits_{k=1}^{K}r_{k}=1$, one can easily get the parameters updating by:
\begin{equation}\label{eq:paras_up}
\left\{
\begin{array}{ll}
r^{\nu+1}_{k}&=\displaystyle\biggl.\frac{\sum\limits_{x\in\Omega}w^{\nu}_{k}(x)}{|\Omega|},\\
(\sigma_{k}^{2})^{\nu+1}&=\displaystyle\biggl.\frac{\sum\limits_{x\in\Omega}(u(x)-f(x))^{2}w^{\nu}_{k(x)}}{\sum\limits_{x\in\Omega}w^{\nu}_{k}(x)}.
\end{array}
\right.
\end{equation}
\textbf{$w$ problem (\ref{eq:w_pro}):}\\
$\mathbf{w}$=$(w_{1},w_{2},...,w_{K})$ subproblem would also have a closed-form solution. The related minimization problem becomes
 \begin{equation*}\label{Hformula_s1}
\begin{array}{ll}
\mathbf{w}^{\nu+1}=\arg\min\limits_{\mathbf{w}\in\mathbb{S}}\displaystyle\biggl.\left\{\frac{1}{2}\sum_{x\in\Omega}
\sum_{k=1}^{K}\frac{\left(u^{\nu+1}(x)-f(x)\right)^2}{(\sigma_k^2)^{\nu+1}}w_k(x)\right.\\
-\displaystyle\biggl.\sum_{x\in\Omega}\sum_{k=1}^{K}w_k(x)\ln r^{\nu+1}_k
+\displaystyle\biggl.\frac{1}{2}\sum_{x\in\Omega}\sum_{k=1}^{K}w_k(x)\ln (\sigma_k^2)^{(\nu+1)}\\
\left.+\displaystyle\biggl.\sum_{x\in\Omega}\sum_{k=1}^{K}w_k(x)\ln w_k(x)\right\}.
\end{array}
\end{equation*}
It has a closed-form solution % (details in Appendix B):
\begin{equation}\label{up_w}
w^{\nu+1}_{k}(x)=\displaystyle\biggl.\frac{\frac{r^{\nu+1}_{k}}{\sigma^{\nu+1}_{k}}e^\frac{{-(u^{\nu+1}(x)-f(x))^{2}}}{2(\sigma^{2}_{k})^{\nu+1}}}{\sum\limits_{k}\frac{r^{\nu+1}_{k}}{\sigma^{\nu+1}_{k}}e^\frac{{-(u^{\nu+1}(x)-f(x))^{2}}}{2(\sigma^{2}_{k})^{\nu+1}}}.
\end{equation}

\begin{figure}
{\includegraphics[width=0.5\textwidth]{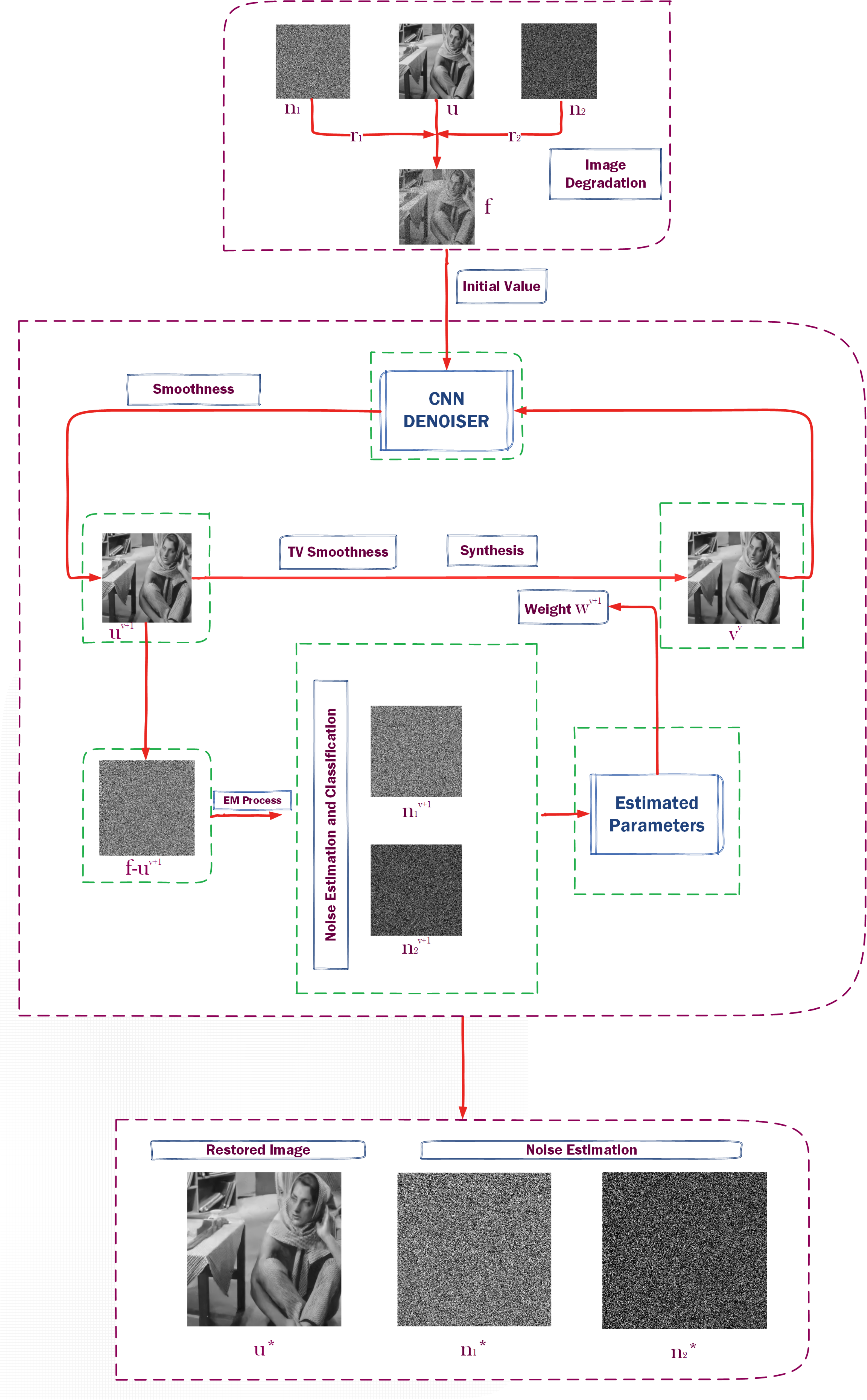}}%{images/liucheng.jpg}}
\caption{The proposed CNN-EM algorithm structure.}
\label{fig:liucheng}
\end{figure}

To sum up, we summarize the algorithm overview of model (\ref{model1}) in algorithm \ref{alg1}.
\begin{algorithm}
	\caption{CNN-EM mixed noise removal}\label{alg1}
Given tolerant error = $\zeta$; Set parameters $\eta=0.8,~\epsilon=1e-2,$
%Set initial value $v^{0},\mu^{0}, r_{k}^{0},(\bf\sigma^{2})^{0}, w_{k}^{0}$,
Set initial value $v^{0}=f,~\mu^{0}=0,~r_{k}^{0}=\frac{1}{K},~\sigma_{1}^{2}=\frac{500}{255^{2}},~\sigma_{2}^{2}=\frac{50}{255^2},~w_{k}^{0}=\frac{1}{K}$,
let $\nu=0$. Do the following steps:\\
\\
1.Smoothness: removing noise by CNN denoiser (\ref{cnn});\\
2.Synthesis: selecting suitable fidelity term for each pixels according to the estimated weighting function by variational step (\ref{up_v}) (\ref{up_d});\\
3.Noise put back: updating the dual variable by (\ref{eq:mu_pro});\\
4.Noise parameter estimation: updating the parameters by (\ref{eq:paras_up});\\
5.Noise Classification: updating the weighting function by (\ref{up_w});\\
\\
6.Convergence checking: if $\mathnormal{\frac{||u^{\nu+1}-u^{\nu}||^{2}}{||u^{\nu}||^{2}}}<\zeta$, stop; else, return to step 1.
\end{algorithm}
The structure of our CNN-EM algorithm is also contained in \figurename\ref{fig:liucheng}.
%\subsection{Impulse Noise}

\subsection{Gaussian Noise Plus Impulse Noise}

As for this part, we assume that the noise model is followed by mixed distribution with Gaussian noise $n_{1}$ and impulse noise $n_{2}$, the noise model \cite{Liu2013} can be written as
\begin{equation}\label{eq:gaussian-impulse}
n=\left\{
\begin{array}{ll}
n_{1},~with~probability~1-r,\\
n_{2},~with~probability~r,\\
\end{array}
\right.
\end{equation}
where $n_{1}$ and $n_{2}$ are the Gaussian noise and uniformly distributed random value range of [0,~255] named random-valued noise or either 0 or 255 named salt-and pepper noise, respectively. In such a case, one can get
\begin{pro}[\cite{Liu2013}]\label{pro1}
The PDF of Gaussian plus random-valued noise and Gaussian plus salt-and-pepper noise have the following expression respectively,
\begin{equation}\label{eq:gaussian-impulse}
p(x)=\left\{
\begin{array}{ll}
(1-r)p_{1}(x)+\frac{r}{255}\int_{-x}^{255-x}p_{2}(y)dy,\\
(1-r)p_{1}(x)+\frac{r}{2}p_{2}(-x)+\frac{r}{2}p_{2}(255-x),\\
\end{array}
\right.
\end{equation}
where $p_{1}$ is a gaussian function and $p_{2}$ is the PDF of clean image with intensity of range [0,~255], which is always expressed by normalized histogram of the clean image.
\end{pro}

Since one can use median filters to well detect salt-and-pepper noise, so some existing works such as two-phase method \cite{Cai2008} \cite{Cai2009} can restore the image well even when the density of noise is as high as 90\%.
However, the random-valued noise is not easy to detected and here we pay more attention on random-valued noise.

In fact, the PDF of random-valued noise can be expressed as
\begin{equation}\label{eq:gaussian-impulse2}
p(x)=\left\{
\begin{array}{ll}
(1-r)p_{1}(x)+\frac{r(255+x)}{255^2},&-255\leq x \leq 0,\\
(1-r)p_{1}(x)+\frac{r(255-x)}{255^2},&0\leq x \leq 255,\\
(1-r)p_{1}(x),&else,
\end{array}
\right.
\end{equation}
if we suppose that the clean image has a normalized histogram, namely $p_2$ is an uniformly distributed PDF in $[0,255]$.
As discussed above, one can use this PDF to construct the data fidelity to complete the model. However, the second part of (\ref{eq:gaussian-impulse2}) is not differential which is hard to optimize. As discussed in the \cite{Liu2013}, in fact, this part can be well approximated by a Gaussian function, which means the model with Gaussian plus impulse noise can be optimized by the model of mixed Gaussian noise.

\section{Experimental Results}

\begin{figure*}
\includegraphics[width=1.0\textwidth]{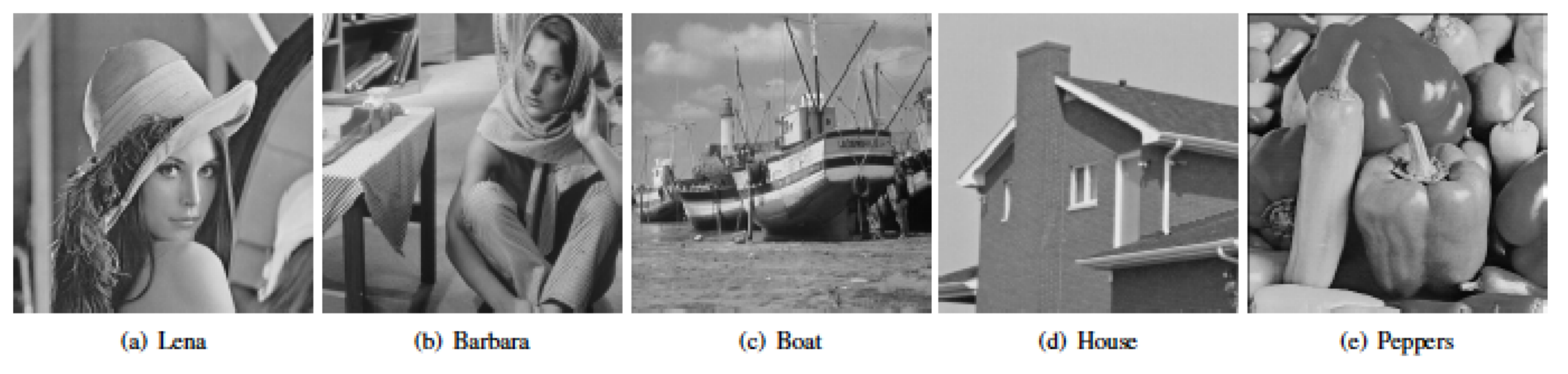}
%\subfigure[Lena]{\includegraphics[width=0.19\textwidth]{images/lena.png}}
%\subfigure[Barbara]{\includegraphics[width=0.19\textwidth]{images/barbara.png}}
%\subfigure[Boat]{\includegraphics[width=0.19\textwidth]{images/boat.png}}
%\subfigure[House]{\includegraphics[width=0.19\textwidth]{images/house.png}}
%\subfigure[Peppers]{\includegraphics[width=0.19\textwidth]{images/peppers.png}}
\caption{Original test images: (a) Lena. (b) Barbara. (c) Boat. (d) House. (e) Peppers.}
  \label{fig:1}
\end{figure*}

In this section, we make comparison between our proposed CNN based regularization model and some related model. Here we give 5 test images in \figurename \ref{fig:1}
 uesd in our experiments: Lena ($512~\times~512$), Barbara ($512~\times~512$), Boat ($512~\times~512$), House ($256~\times~256$) and Peppers ($256~\times~256$). To estimate the denoising quality of the different methods, we adopt PSNR value
 \begin{equation*}
 \text{PSNR}=\frac{255^{2}}{Var(u^{*}-u)}
 \end{equation*}
 as the quality index.
 Here $u$ and $u^{*}$ are the clean and denoised image, respectively.

\subsection{Gaussian Mixed Noise}
 In the first experiment, we give the restored results under mixed Gaussian noise. To make comparison, we take K-SVD method {\cite{Aharon2006}}, W-KSVD model\cite{Liu2013}, and CNN denoising method \cite{Zhang2017learning} as reference.

\begin{figure*}
\includegraphics[width=1.0\textwidth]{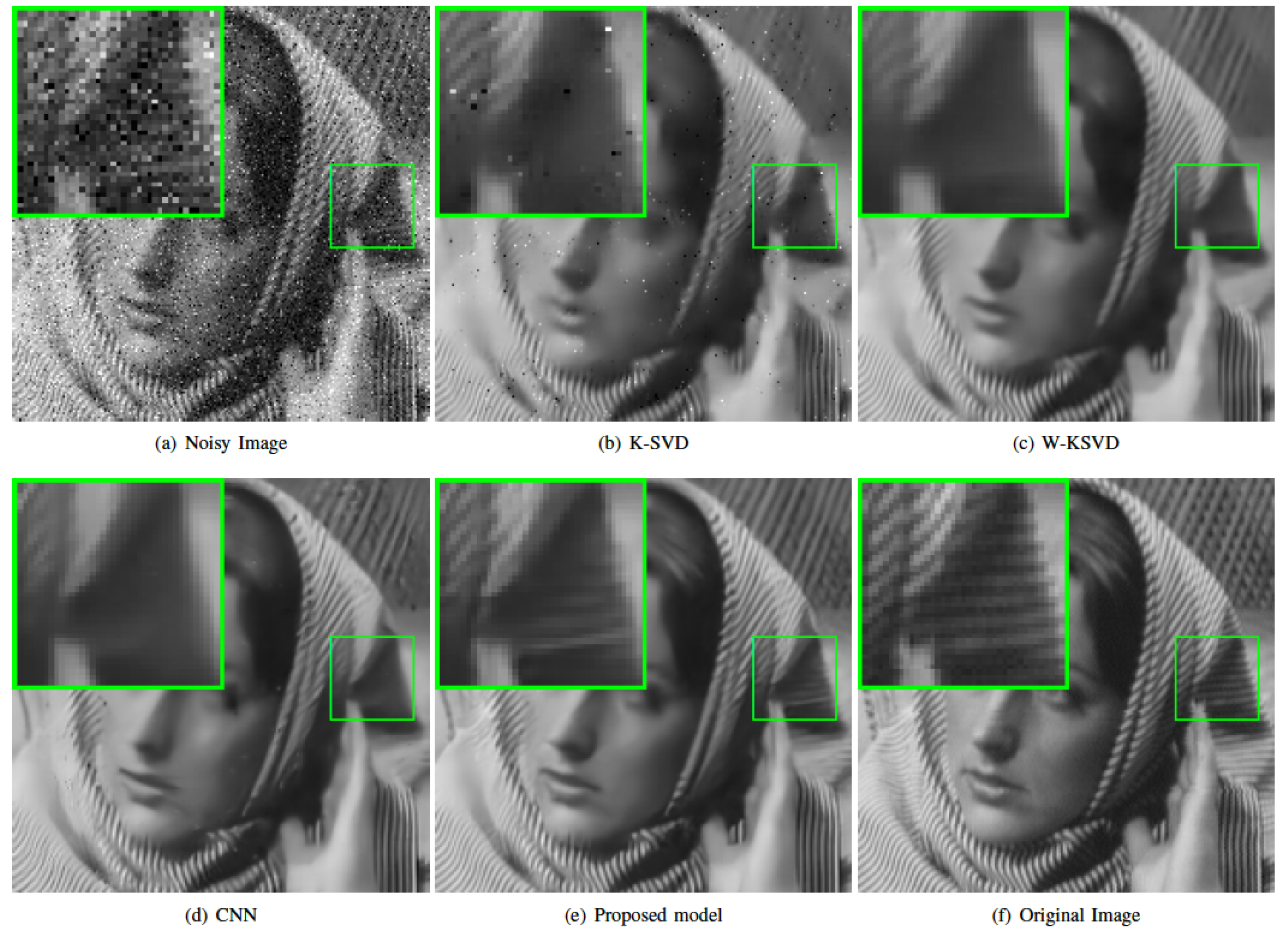}
%\subfigure[Noisy Image]{\includegraphics[width=0.33\textwidth]{images/barbara_noisy_part2.png}}
%\subfigure[K-SVD]{\includegraphics[width=0.33\textwidth]{images/barbara_ksvd_part2.png}}
%\subfigure[W-KSVD]{\includegraphics[width=0.33\textwidth]{images/barbara_wksvd_part2.png}}\\
%\subfigure[CNN]{\includegraphics[width=0.33\textwidth]{images/barbara_cnn_part2.png}}
%\subfigure[Proposed model]{\includegraphics[width=0.33\textwidth]{images/barbara_prop_part2.png}}
%\subfigure[Original Image]{\includegraphics[width=0.33\textwidth]{images/barbara_ori_part.png}}
\caption{Results under mixed Gaussian ($r_{1}:r_{2}=0.7:0.3,~\sigma_{1}=10,~\sigma_{2}=50$): (a) Noisy "Barbara"(PSNR=19.02). (b) K-SVD ({\cite{Aharon2006}} with known parameters, PSNR=26.95). (c) W-KSVD (\cite{Liu2013} with unknown parameters, PSNR=30.07). (d) CNN (\cite{Zhang2017learning}, PSNR=28.95). (e) Ours (PSNR=30.69). (f) Original clean image.}
  \label{fig:3}
\end{figure*}

The test image ``Barbara" is corrupted by mixed Gaussian noise with mixture ratio $r_{1}:r_{2}=0.7:0.3$ and the standard deviations $\sigma_{1}=10$ and $\sigma_{2}=50$, respectively. Though the K-SVD model is design for single Gaussian distribution, we still list the results as reference to show the superiority of EM parameter estimation. The noise variances appeared in K-SVD method are set as  $r_{1}\sigma^{2}_{1}+r_{2}\sigma^{2}_{2}$ according to proposition 5 in the work \cite{Liu2013}. For W-KSVD \cite{Liu2013} which integrating weighted dictionary learning and sparse coding based on Expectation Maximum(EM) process used for mixed noise removal, we update all the parameters, including weights parameter $r_{k}$ and variance $\sigma_{k}$. To compare with the learning based methods, we give the results by the latest work in CVPR2017 \cite{Zhang2017learning} called ``CNN based method" here which is trained to remove additive Gaussian noise with level range $[0,50]$. We show the noisy image and the corresponding restored results in \figurename \ref{fig:3}. We zoom in the regions in green rectangle which is placed in the left-bottom of each image patch.

To be compared with the K-SVD methods, one can find that some speckles exist in the results by K-SVD method \figurename \ref{fig:3}(b) since K-SVD can not distinguish the different noise levels, and our proposed variational model \figurename \ref{fig:3}(e) which is based on EM process with parameters updating can preciously determine the noise level of each pixels. To compare with the W-KSVD methods \figurename \ref{fig:3}(c), our proposed model can preserve image detail preciously, such as texture information, since our proposed model has high level image prior so as to having better denoising performance. CNN method performs well in denoising process, however, since
the noise is not a standard Gaussian distribution and we do not have the exact noise variance, if the initial value of noise level are far from the real noise level in CNN process, the restored image will be very  bad. To be contrasted, our model are separated to four step including noise level estimation and image synthesis process by operator splitting, the estimation of noise can endow CNN process a better noise level estimation to have a better denoising performance \figurename \ref{fig:3}(e). Meanwhile, the CNN process largely depends on the labeled samples, if the noise distribution or noise level are not included in the sample database, the restored image are always undesirable. The image synthesis process can partly release the sample-dependent effect so as to performing robust. Moreover, our proposed model has a high efficiency, \tablename~\ref{table_1}  shows the CPU run times for denoising the images with mixed noise ($r_{1}:r_{2} = 0.7:0.3,~\sigma_{1}=5,~\sigma_{2}=30$) of size 256$\times$256, 512$\times$512 by different methods, including K-SVD {\cite{Aharon2006}}, W-KSVD {\cite{Liu2013}}, CNN {\cite{Zhang2017learning}} and proposed algorithm, here the time of W-KSVD and proposed model shown in \tablename~\ref{table_1} are the time for one outer iteration by each method. It can be seen that the proposed CNN-EM algorithm is more than 30 times faster than W-KSVD \cite{Liu2013}.

\begin{table}
\centering
\caption{Run Time of Different Methods: K-SVD {\cite{Aharon2006}}, W-KSVD {\cite{Liu2013}}, CNN {\cite{Zhang2017learning}} and Proposed Algorithm.}\label{table_1}
%\resizebox{\textwidth}{28mm}
{
\begin{tabular}{c|c|c|c|c}
\hline
Size&K-SVD\cite{Aharon2006}&W-KSVD\cite{Liu2013}&CNN\cite{Zhang2017learning}&Proposed\\
\hline
$256\times256$&44.999s&492.28s/10&1.28s&13.44s/10\\
\hline
$512\times512$&128.351s&1245.70s/10&3.68s&38.09s/10\\
\hline
\end{tabular}}
\end{table}

\begin{figure*}
\includegraphics[width=1.0\textwidth]{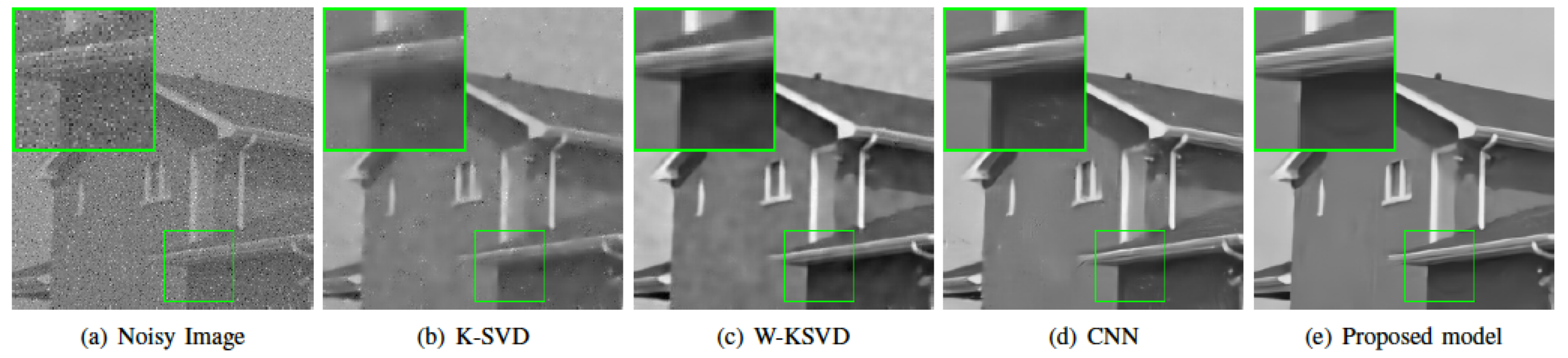}
%\subfigure[Noisy Image]{\includegraphics[width=0.19\textwidth]{images/rreexp1_noisy_house.png}}
%\subfigure[K-SVD]{\includegraphics[width=0.19\textwidth]{images/rreexp1_ksvd_house.png}}
%\subfigure[W-KSVD]{\includegraphics[width=0.19\textwidth]{images/rreexp1_wksvd_house.png}}
%\subfigure[CNN]{\includegraphics[width=0.19\textwidth]{images/rreexp1_cnn_house.png}}
%\subfigure[Proposed model]{\includegraphics[width=0.19\textwidth]{images/rreexp1_prop_house.png}}
\caption{Results under mixed Gaussian ($r_{1}:r_{2}=0.7:0.3,~\sigma_{1}=15,~\sigma_{2}=75$): (a) Noisy "House" (PSNR=15.41). (b) K-SVD ({\cite{Aharon2006}} with known parameters, PSNR=27.41). (c) W-KSVD (\cite{Liu2013} with unknown parameters, PSNR=31.42). (d) CNN (\cite{Zhang2017learning}, PSNR=30.57). (e) Ours (PSNR=32.77).}
  \label{fig:exp1_1}
\end{figure*}

\begin{figure*}
\includegraphics[width=1.0\textwidth]{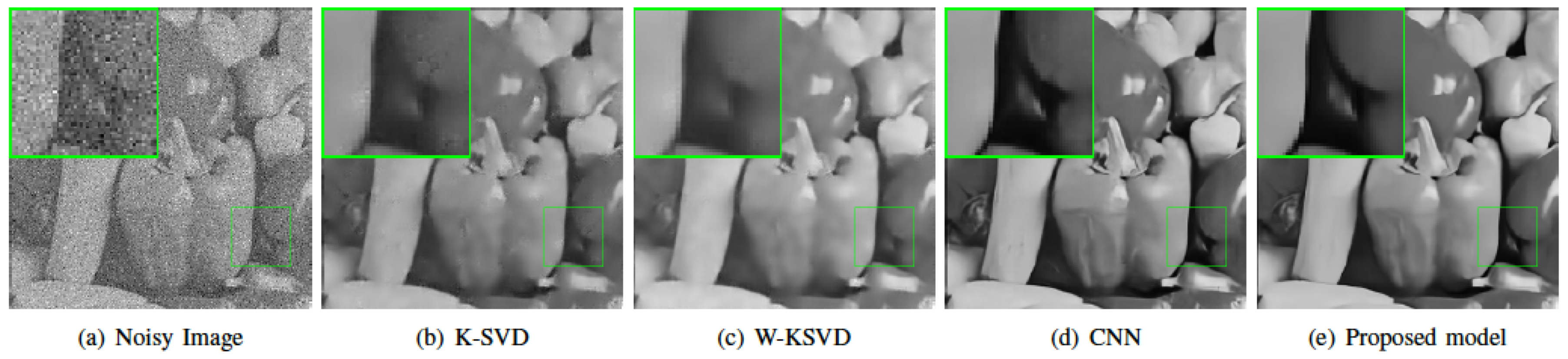}
%\subfigure[Noisy Image]{\includegraphics[width=0.19\textwidth]{images/reexp1_noisy_peppers256.png}}
%\subfigure[K-SVD]{\includegraphics[width=0.19\textwidth]{images/reexp1_ksvd_peppers256.png}}
%\subfigure[W-KSVD]{\includegraphics[width=0.19\textwidth]{images/reexp1_wksvd_peppers256.png}}
%\subfigure[CNN]{\includegraphics[width=0.19\textwidth]{images/reexp1_cnn_peppers256.png}}
%\subfigure[Proposed model]{\includegraphics[width=0.19\textwidth]{images/reexp1_prop_peppers256.png}}
\caption{Results under mixed Gaussian ($r_{1}:r_{2}=0.3:0.7,~\sigma_{1}=10,~\sigma_{2}=50$): (a) Noisy "Peppers" (PSNR=15.57). (b) K-SVD ({\cite{Aharon2006}} with known parameters, PSNR=26.82). (c) W-KSVD (\cite{Liu2013} with unknown parameters, PSNR=27.37). (d) CNN (\cite{Zhang2017learning}, PSNR=28.16). (e) Ours (PSNR=28.79).}
  \label{fig:exp1_2}
\end{figure*}

Here we give another two numerical experiments, the images used for algorithm comparisons are corrupted by mixed noise. And the sample image ``House" is contaminated by mixed Gaussian $r_{1}:r_{2}=0.7:0.3,~\sigma_{1}=15,~\sigma_{2}=75$, while ``Peppers" is contaminated by mixed Gaussian $r_{1}:r_{2}=0.3:0.7,~\sigma_{1}=10,~\sigma_{2}=50$. \figurename\ref{fig:exp1_1} and \figurename\ref{fig:exp1_2} show the restored images by each methods corresponding to the artificial test images. From the restored images by each methods, one can easily get the same conclusion as the last experiment \figurename\ref{fig:1}. We test our method in every sample images, and list all the PSNR values of results by different methods in \tablename~\ref{table_2}. It can be found that almost all the restored images by proposed model have the highest PSNR values, which shows the superiority of our model. Here we pay more attention on the noise mixture $r_{1}:r_{2} = 0.3:0.7$ and $\sigma_{1}=15,~\sigma_{2}=75$. As discussion above, the noise level to be set in CNN denoiser is $\sqrt{r_{1}\sigma^{2}_{1}+r_{2}\sigma^{2}_{2}}=63.2851$, which is out of the range [0, 50] of the denoisers \cite{Zhang2017learning}. Under this situation, the CNN denoiser is of invalidation in fact. However, since the existence of the step of image synthesis with related to $v$ subproblem, our proposed model can behave better and more robust than the state-of-art methods, which can be seen as modification of the CNN denoiser and also shows the superiority of our proposed model.


\begin{table*}
\centering
\caption{PSNR Values: K-SVD {\cite{Aharon2006}}, W-KSVD {\cite{Liu2013}}, CNN {\cite{Zhang2017learning}} and Proposed Algorithm.}\label{table_2}
%\resizebox{\textwidth}{28mm}
{
\begin{tabular}{ccccccccccccc}
\hline
Images & & $\sigma_{1}$=5 & $\sigma_{2}=30$ & & &$\sigma_{1}$=10 & $\sigma_{2}=50$ & & &$\sigma_{1}$=15 & $\sigma_{2}=75$ & \\
\cline{3-5}\cline{7-9}\cline{11-13}
$\downarrow$ &$r_{1}:r_{2}\rightarrow$&0.3:0.7&0.5:0.5&0.7:0.3&&0.3:0.7&0.5:0.5&0.7:0.3&&0.3:0.7&0.5:0.5&0.7:0.3\\
\hline
\hline
Lena&K-SVD \cite{Aharon2006}&31.11&31.60&31.30& &28.49&28.92&28.43& &26.41&26.79&26.05\\
    &W-KSVD \cite{Liu2013}&31.43&32.69&34.24& &29.00&30.43&32.07& &27.04&28.57&30.36\\
    &CNN \cite{Zhang2017learning} &32.34&32.93&33.72& &30.20&30.79&31.56& &19.55&27.05&29.83\\
    &Proposed&\textbf{32.88}&\textbf{34.18}&\textbf{35.55}& &\textbf{30.86}&\textbf{32.12}&\textbf{33.50}& &\textbf{29.04}&\textbf{30.36}&\textbf{32.12}\\
    %&Proposed&32.88&34.18&35.55&30.86&32.12&33.50&29.04&30.36&32.12\\
\hline
Barbara&K-SVD \cite{Aharon2006}&29.37&30.08&30.65&&26.40&27.08&26.95&&23.70&24.38&24.21\\
       &W-KSVD \cite{Liu2013}&29.19&30.50&32.69&&26.46&28.15&30.07&&23.75&26.20&28.22\\
       &CNN \cite{Zhang2017learning} &\textbf{29.80}&30.57&31.66&&27.15&27.90&28.95&&19.22&24.60&26.81\\
       &Proposed&{29.76}&\textbf{31.35}&\textbf{32.87}&&\textbf{27.67}&\textbf{28.88}&\textbf{30.69}&&\textbf{25.23}&\textbf{26.68}&\textbf{28.94}\\
     %  &Proposed&29.76&31.35&32.87&27.67&28.88&30.69&25.23&26.68&28.94\\
\hline
Boat&K-SVD\cite{Aharon2006}&29.08&29.63&29.78&&26.20&27.07&26.81&&24.61&25.02&24.60\\
    &W-KSVD \cite{Liu2013}&28.96&29.85&31.19&&26.79&27.82&29.19&&25.02&26.31&27.77\\
    &CNN \cite{Zhang2017learning} &29.96&30.60&31.50&&27.83&28.43&29.22&&19.28&25.43&27.52\\
    &Proposed&\textbf{29.99}&\textbf{31.66}&\textbf{33.14}&&\textbf{28.29}&\textbf{29.43}&\textbf{30.92}&&\textbf{26.34}&\textbf{27.65}&\textbf{29.39}\\
    %&Proposed&29.99&31.66&33.14&28.29&29.43&30.92&26.34&27.65&29.39\\
\hline
House&K-SVD \cite{Aharon2006}&31.81&32.25&31.74&&28.83&29.42&28.88&&26.13&26.74&27.41\\
     &W-KSVD \cite{Liu2013} &32.73&33.66&34.83&&30.16&31.56&33.10&&27.24&29.44&31.42\\
     &CNN \cite{Zhang2017learning} &33.07&33.54&34.26&&30.89&31.56&32.32&&19.33&27.30&30.57\\
     &Proposed&\textbf{33.59}&\textbf{34.80}&\textbf{36.32}&&\textbf{31.56}&\textbf{32.86}&\textbf{34.10}&&\textbf{29.68}&\textbf{30.99}&\textbf{32.77}\\
     %&Proposed&33.59&34.80&36.32&31.56&32.86&34.10&29.68&30.99&32.77\\
\hline
Peppers&K-SVD \cite{Aharon2006}&29.47&30.10&30.32&&26.82&27.40&27.10&&24.53&25.16&24.69\\
       &W-KSVD \cite{Liu2013} &29.66&30.55&31.69&&27.37&28.47&29.79&&25.21&26.76&28.37\\
       &CNN \cite{Zhang2017learning} &30.62&31.38&32.35&&28.16&28.86&29.87&&19.08&25.34&27.88\\
       &Proposed&\textbf{31.18}&\textbf{32.42}&\textbf{33.89}&&\textbf{28.79}&\textbf{30.10}&\textbf{31.69}&&\textbf{26.91}&\textbf{28.22}&\textbf{30.00}\\
      % &Proposed&31.18&32.42&33.89&28.79&30.10&31.69&26.91&28.22&30.00\\

\hline
\end{tabular}}
\end{table*}

Since the high calculation efficiency of our proposed algorithm, which can be tested on the image data-set, we make comparison between CNN based method {\cite{Zhang2017learning}} and our proposed algorithm on BSDS500 dataset \cite{amfm_pami2011}. \tablename~\ref{table_3} gives the contrast results on 100 images, 300 images and total 500 images of BSDS500 dataset. One can find our proposed algorithm have higher PSNR value. Our proposed model has at least 0.31 db improvement and 1.61db improvement for average than original CNN based method {\cite{Zhang2017learning}}.

\begin{table*}
\centering
\caption{PSNR Values: CNN {\cite{Zhang2017learning}} and Proposed Algorithm on BSDS500 DataSet \cite{amfm_pami2011}.}\label{table_3}
%\resizebox{\textwidth}{28mm}
{
\begin{tabular}{ccccccccccccc}
\hline
DataSet & & $\sigma_{1}$=5 & $\sigma_{2}=30$ & & &$\sigma_{1}$=10 & $\sigma_{2}=50$ & & &$\sigma_{1}$=15 & $\sigma_{2}=75$ & \\
\cline{3-5}\cline{7-9}\cline{11-13}
$\downarrow$ &$r_{1}:r_{2}\rightarrow$&0.3:0.7&0.5:0.5&0.7:0.3&&0.3:0.7&0.5:0.5&0.7:0.3&&0.3:0.7&0.5:0.5&0.7:0.3\\
\hline
\hline
BSDS500 \cite{amfm_pami2011}    &CNN \cite{Zhang2017learning} &29.27&29.99&30.97& &27.08&27.68&28.47& &19.17&24.76&26.77\\
100 images      &Proposed&\textbf{29.60}&\textbf{30.73}&\textbf{32.20}& &\textbf{27.46}&\textbf{28.42}&\textbf{29.81}& &\textbf{25.74}&\textbf{26.85}&\textbf{28.27}\\
\hline
BSDS500 \cite{amfm_pami2011}    &CNN \cite{Zhang2017learning} &29.28&30.00&31.00& &27.06&27.67&28.49& &19.22&24.76&26.75\\
300 images      &Proposed&\textbf{29.58}&\textbf{30.68}&\textbf{32.15}& &\textbf{27.40}&\textbf{28.35}&\textbf{29.76}& &\textbf{25.67}&\textbf{26.76}&\textbf{28.20}\\
\hline
BSDS500 \cite{amfm_pami2011}    &CNN \cite{Zhang2017learning} &29.21&29.92&30.91& &27.03&27.62&28.42& &19.23&24.76&26.71\\
500 images      &Proposed&\textbf{29.51}&\textbf{30.61}&\textbf{32.06}& &\textbf{27.37}&\textbf{28.30}&\textbf{29.69}& &\textbf{25.68}&\textbf{26.74}&\textbf{28.16}\\
\hline
\end{tabular}}
\end{table*}

\begin{figure}
%\subfigure[]
{\includegraphics[width=0.5\textwidth]{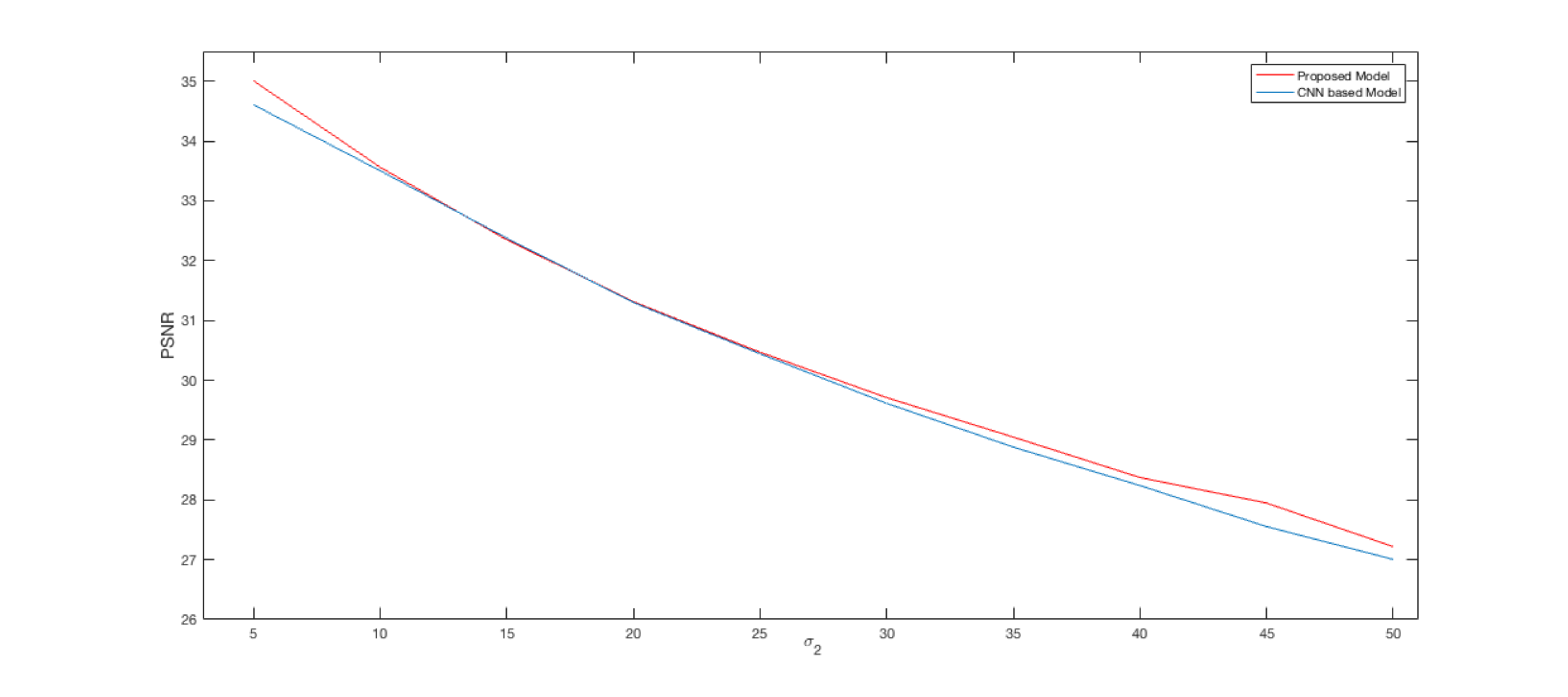}}
%\subfigure[]{\includegraphics[width=0.5\textwidth]{images/r1.png}}
  \caption{PSNR values for CNN \cite{Zhang2017learning} and the proposed CNN-EM
  under 2-component mixed Gaussian noise with fixed $\sigma_1$ and increasing $\sigma_{2}=5:5:50$. The mixed ratio is fixed as $r_{1}:r_{2}=0.3:0.7$.}
  \label{fig:6}
\end{figure}

In the next experiment, we explore the relationship between PSNR values and the noise level of mixed Gaussian noise with sample image ``Barbara" by proposed model and CNN based model \cite{Zhang2017learning}. Here, we set noise ratio $r_{1}:r_{2}$ = $0.3:0.7$, and one of the noise with fixed standard deviations $\sigma_{1}=15$ and the other noise with increasing noise level $\sigma_{2}$ from 5 to 50 with step size 5 and give the results in \figurename \ref{fig:6}. One can find that the PSNR values of both methods are decreasing with the increasing of $\sigma_{2}$ as we expected. Meanwhile, the PSNR values nearby the value $\sigma_{2}=\sigma_{1}$ are get closer between CNN based method and proposed method, since if $\sigma_{1}=\sigma_{2}$, the mixed noise model will degenerate to single Gaussian noise model, which can be solved by CNN based model efficiently. with high noise level which is far from $\sigma_{1}=15$, our proposed model has more satisfactory behaviours.

\begin{figure}
%\subfigure[]{\includegraphics[width=0.5\textwidth]{images/sigma2.png}}
%\subfigure[]
{\includegraphics[width=0.5\textwidth]{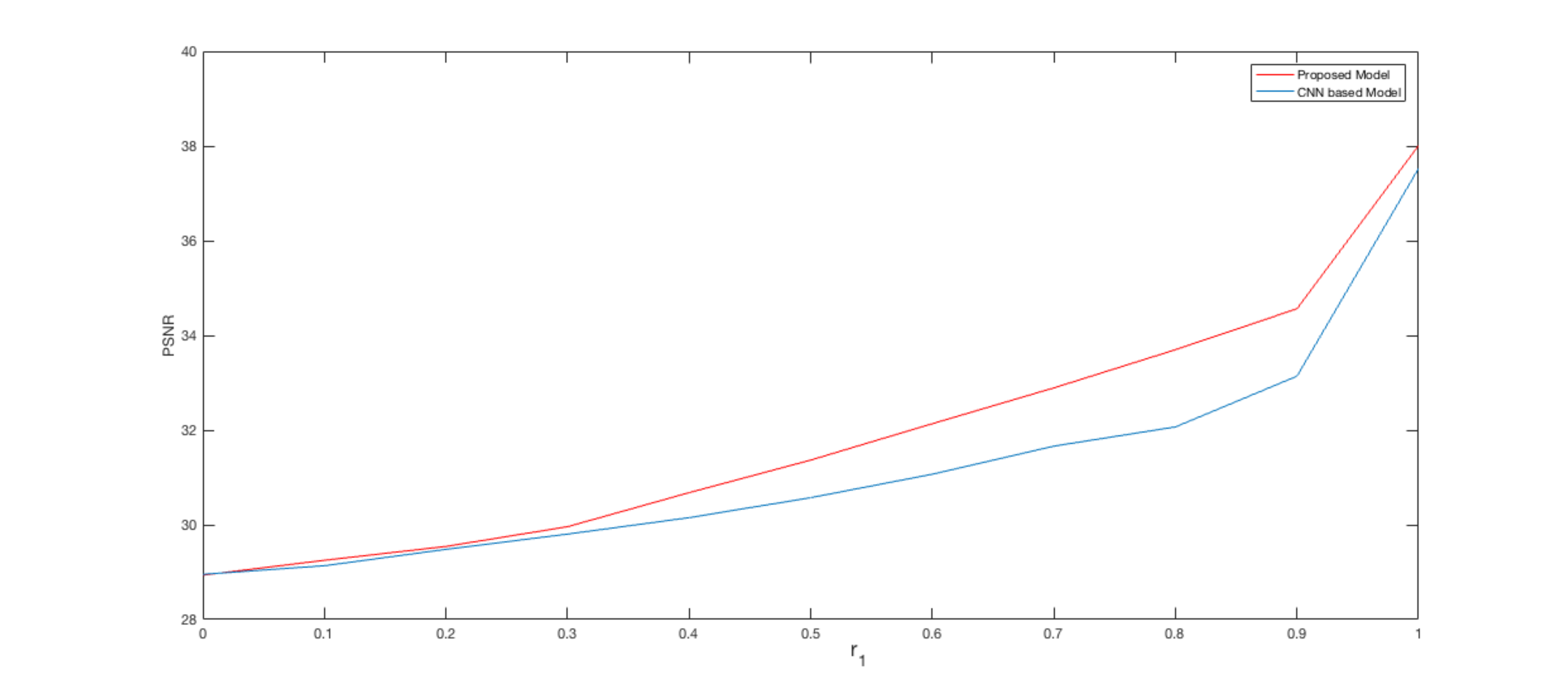}}
  \caption{
  PSNR values for CNN \cite{Zhang2017learning} and the proposed CNN-EM
  under 2-component mixed Gaussian noise with increasing
  mixed ratio $r_{1}=0:0.1:1$ and $r_{2}=1-r_{1}$. The noise variances $\sigma_{1}=5$, $\sigma_{2}=30$ are fixed.}
  \label{fig:7}
\end{figure}

In \figurename \ref{fig:7}, we give relationship between PSNR values and the mixed ratio of mixed Gaussian noise with sample image ``Barbara", and $\sigma_{1}=5,~\sigma_{2}=30$ and mixed ratio: $r_{1}$ $=$ $0:0.1:1$ and $r_{2}$ $=$ $1-r_{1}$. Here we also show the results of the CNN based method \cite{Zhang2017learning} which serves as contrast. In fact, with the increasing of the $r_{1}$, the valid noise level $\sqrt{r_{1}\sigma_{1}^{2}+r_{2}\sigma_{2}}$ is decreasing, which dues to the increasing of the PSNR values by both methods. Otherwise, with small $r_{1}$ ($r_{1}$ is close to 0) or large $r_{1}$ (close to 1, which means $r_{2}$ is closer to 0), the PSNR value by CNN based method \cite{Zhang2017learning} and our proposed model are close to each other, since at this time, the noise in fact can be seen as single Gaussian noise, which meets our expectation.

%\subsection{Impulse Noise}
%
%Commonly, there are two types of impulse noise: salt-and-pepper noise and random-value noise. Though impulse noise are not strictly additive noise, however numerical experiments shows that our proposed model can work well and behave better than the state-of-art impulse denoising methods.

\subsection{Gaussian Noise Plus Impulse Noise}

In fact, our model also can work on the images with Gaussian noise plus Impulse noise, here we test our model on ``Barbara" contaminated by Gaussian-Random Value noise, and the density of ranom-value noise is set $r = 0.3$, and the standard deviation of Gaussian is set $\sigma = 15$. To obtain better restoration of noisy image ``Barbara", here we set the initial $w^{0}$ as the output of the first phase of two-phase models \cite{Cai2008} \cite{Cai2009} \cite{Liu2013} as
\begin{equation}\label{eq:gaussian-impulse3}
w_{i}^{0}=\left\{
\begin{array}{ll}
1,&~if~f_{i}==u^{med}_{i},\\
0,&~else,
\end{array}
\right.
\end{equation}
where $u^{med}$ is the result of the median filter. Here we detect random-valued impulse noise by adaptive center-weighted median filter
(ACWMF) \cite{Ko1991}. Meanwhile, we set the initial variances for impulse noise as \cite{Liu2013}
\begin{equation}\label{eq:sigma2}
\left\{
\begin{array}{ll}
(\sigma_{1}^{2})^{0}=\frac{\sigma^{2}}{10},\\
(\sigma_{2}^{2})^{0}=\frac{9\sigma^{2}}{10},\\
\end{array}
\right.
\end{equation}
where $\sigma^{2}$ can be estimated by following mode \cite{Liu2013}
$$
\sigma^{2}\approx \frac{\sum\limits_{i=1}^{N_{1}}\sum\limits_{j=1}^{N_{2}}\Delta f_{ij}}{10N_{1}N_{2}}.
$$

\begin{figure*}
\includegraphics[width=1.0\textwidth]{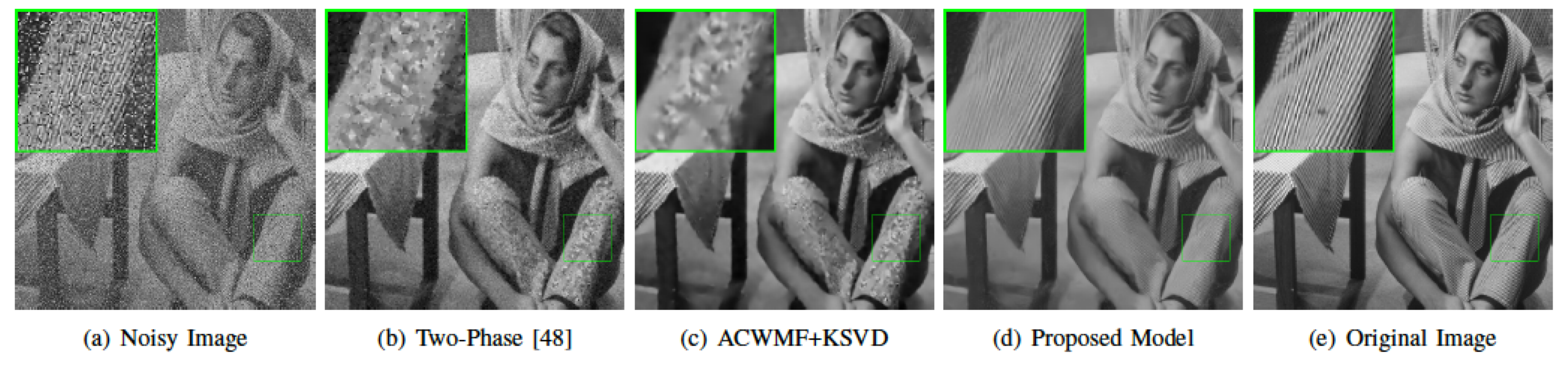}
%\subfigure[Noisy Image]{\includegraphics[width=0.19\textwidth]{images/reexp2_prop_noise_barbara.png}}
%\subfigure[Two-Phase\cite{Cai2009}]{\includegraphics[width=0.19\textwidth]{images/reexp2_twophase_3_3_barbara.png}}
%\subfigure[ACWMF+KSVD]{\includegraphics[width=0.19\textwidth]{images/reexp2_acwmf_ksvd_3_3_barbara.png}}
%\subfigure[Proposed Model]{\includegraphics[width=0.19\textwidth]{images/reexp2_prop_3_3_barbara.png}}
%\subfigure[Original Image]{\includegraphics[width=0.19\textwidth]{images/barbara_ori_3_3(1).png}}
\caption{Results and the Zooming in region in green rectangle with Barbara contaminated by Gaussian plus random value: (a) Noisy "Barbara" ($r=0.3,~\sigma=15$, PSNR=13.81). (b) Two-Phase (\cite{Cai2009} with known parameters, PSNR=22.67). (c) ACWMF+KSVD \cite{Ko1991}\cite{Aharon2006} with unknown parameters, PSNR=23.77). (d) Ours (PSNR=24.62). (e)Original clean image.}
  \label{fig:exp2_2}
\end{figure*}

For comparison, we give the results by some related model: ACWMF \cite{Ko1991} plus K-SVD {\cite{Aharon2006}} (first using ACWMF to filter the noisy image, then denoising image by K-SVD), two-phase model \cite{Cai2009} \cite{Xiao2011} which are two good mixed noise removal models. The parameters for these two two-phase models are chosen as with the highest PSNR and noise variance is set as ${r_{1}\sigma_{1}^{2}+r_{2}\sigma_{2}}$ with $\sigma_{1},~\sigma_{2}$ in (\ref{eq:sigma2}). The restored images by Two-Phase, ACWMF+KSVD and our proposed model are shown in \figurename\ref{fig:exp2_2}, and to have a better visualization, we zoom in the region in green rectangle which is shown in the left-bottom of \figurename\ref{fig:exp2_2}. From the restored results, our proposed model has better behaviour with no doubt, especially among the texture regions. Meanwhile, the above mentioned two methods in fact are used for single Gaussian removal, as for mixed noise, these two model can not distinguish the noise level and type, so there will be some speckles inevitably. To be contrasted, our restored image is good enough visually.

In the next experiment, we test our model on more mixed noise combination, here the ``Barbara" is contaminated by different level of Gaussian noise $\sigma=5,~10,~15$ and changing density of random-valued noise $r=0.1,~0.2,~0.3$. The results under Gauusian plus random-valued noise by our model and the related models are shown in TABLE IV. From this table, the results by our proposed model have the highest PSNR, which is coincident with the visual results and also shows the superiority of our model.

%\begin{figure*}
%\subfigure[Noisy Image]{\includegraphics[width=0.22\textwidth]{images/exp2_prop_noise_barbara.png}}
%\subfigure[Two-Phase\cite{Cai2009}]{\includegraphics[width=0.22\textwidth]{images/exp2_twophase_3_3_barbara.png}}
%\subfigure[ACWMF+KSVD]{\includegraphics[width=0.22\textwidth]{images/exp2_acwmf_ksvd_3_3_barbara.png}}
%\subfigure[Proposed Model]{\includegraphics[width=0.22\textwidth]{images/exp2_prop_3_3_barbara.png}}
%%\subfigure[Proposed model]{\includegraphics[width=0.19\textwidth]{images/reexp1_prop_peppers256.png}}
%\caption{Results with Barbara contaminated by Gaussian plus random value: (a) Noisy "Barbara" ($r=0.3,~\sigma=15$, PSNR=13.81). (b) Two-Phase (\cite{Cai2009} with known parameters, PSNR=22.67). (c) ACWMF+KSVD \cite{Ko1991}\cite{Aharon2006} with unknown parameters, PSNR=23.77). (d) Ours (PSNR=24.62).}
%  \label{fig:exp2_1}
%\end{figure*}

\begin{table*}\label{table_4}
\centering
\caption{PSNR Values of Different Methods with Gaussian Noise Plus Random-Valued Noise for Barbara.}
{
\begin{tabular}{ccccccccccccc}
\hline
&  & $\sigma=5$ & & && $\sigma=10$ & & & & $\sigma=15$ & \\
\cline{2-4}\cline{6-8}\cline{10-12}
$r\rightarrow$&0.1&0.2&0.3&&0.1&0.2&0.3&&0.1&0.2&0.3\\
\hline
\hline
Noisy&18.76&15.76&14.04& &18.43&15.61&13.95& &17.94&15.38&13.81\\
Two-phase &25.40&24.77&24.13& &24.34&23.94&23.45& &23.32&23.02&22.67\\
ACWMF+K-SVD  &26.07&25.27&24.51& &25.50&24.91&24.31& &24.64&24.19&23.77\\
$l_1-l_{0}$\cite{Xiao2011} &30.45&27.75&25.95&&28.45&26.59&25.34&&27.33&25.69&24.55\\
%W-KSVD\cite{Liu2013}& 32.97&31.53&28.95&&30.42&28.32&26.30&&28.37&25.98&24.01\\
Proposed&\textbf{30.54}&\textbf{28.75}&\textbf{26.62}& &\textbf{30.89}&\textbf{28.80}&\textbf{25.88}& &\textbf{29.66}&\textbf{27.27}&\textbf{24.62}\\
\hline
\end{tabular}}
\end{table*}

\begin{figure*}
\includegraphics[width=1.0\textwidth]{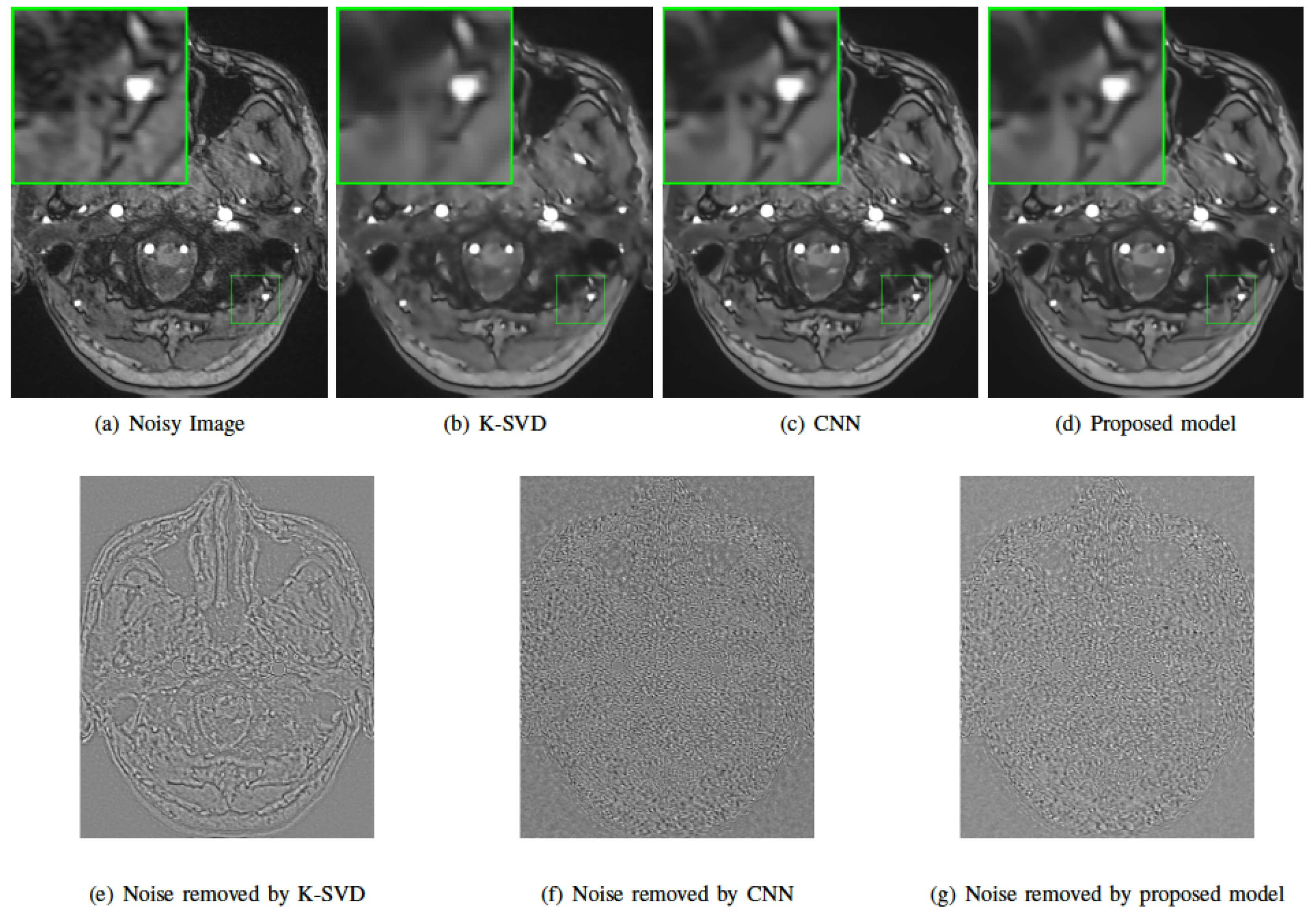}
 \caption{(a)Real noisy image. (b)Restored image by K-SVD {\cite{Aharon2006}}. (c)Restored image by CNN based method \cite{Zhang2017learning}. (d)Restored image by the proposed method. (e)(f)(g)Noise removed by K-SVD, CNN and proposed method, respectively.}
  \label{fig:real}
\end{figure*}

Moreover, we test our proposed model on some real noisy images, the results are shown in \figurename \ref{fig:real} by K-SVD {\cite{Aharon2006}}, CNN based method \cite{Zhang2017learning} and the proposed method, where \figurename \ref{fig:real}(a) is the real noisy brain MR image, \figurename \ref{fig:real}(b), \figurename \ref{fig:real}(c), \figurename \ref{fig:real}(d) are the denoised images by K-SVD, CNN based method and proposed method respectively, the noise removed by K-SVD, CNN, proposed method are shown in \figurename\ref{fig:real}(e), \figurename \ref{fig:real}(f), \figurename \ref{fig:real}(g), where the regions in green rectangle are zoomed in and placed in the left-bottom of the each sub-figures. From the noise removed by different method, we can find that our propose method get better restored results than K-SVD method, since there is less information removed by proposed method. Meanwhile, we find that the restored result by the proposed method is slightly better than CNN.
In this experiment, the noise difference is small, as mentioned before, under this situation, the difference of results provided by CNN and proposed model would be relatively small.

%\subsection{Nonparametric Model}

\section{Conclusion}

We propose a variational mixed noise removal model integrating the CNN deep learning regularization. The variational based fidelity is originated from the EM process treated as the estimation of noise distribution, which can measure the discrepancy between the true value and the observed data accordantly. The CNN based regularization shows better noise removal performance, since CNN can seize more image priori existing in the natural images through training process of large amount of labeled samples. To fill the gap between variational framework and nonlinear CNN regularization, we employ the well-known operator splitting method to separate our model into four parts: noise removal (based on CNN regularization), synthesis, parameters estimation an noise classification, where each step can be optimized efficiently including CNN based denoising since the corresponding subproblem is a standard Gaussian additive model which can be solved by differnet kinds of learning based denoisers.

In fact, parameters estimation and noise classification which come from EM process play a fatal role in the CNN based noise removal step. The EM noise estimation can endow CNN process a better noise level estimation to have a better denoising behaviours. Besides, since the CNN denoiser is  data-dependent, if the noise distribution or noise level is not included in the sample database, the restored image is always undesirable. The image synthesis process can partly release the sample-dependent effect so as to performing robust.

The key point of our model is integrating the CNN regularization into a EM based variational framework, maybe the idea can be used into more extensive image processing field, such as CNN regularization based segmentation and registration.
\section*{Acknowledgment}

Jun Liu and Haiyang Huang were partly supported by The National Key Research and Development Program of China (2017YFA0604903).
\ifCLASSOPTIONcaptionsoff
  \newpage
\fi

%\bibliographystyle{IEEEtran}
%\bibliography{ref}
%\begin{bibliography}

%

\end{document}